\documentclass[letterpaper]{article} 
\usepackage{aaai23}  
\usepackage{times}  
\usepackage{helvet}  
\usepackage{courier}  
\usepackage[hyphens]{url}  
\usepackage{graphicx} 
\usepackage{natbib}  
\usepackage{caption} 
\usepackage{algorithm}
\usepackage{algorithmic}

\usepackage{makecell}
\usepackage{comment}
\usepackage{amsmath,amssymb}
\usepackage{newtxmath}
\usepackage{bbding}
\usepackage{subcaption}
\usepackage{tikz}
\newcommand{\etal}{{et al. }}

\newcommand{\eg}{{e.g.}}

\usepackage{newfloat}
\usepackage{listings}
\DeclareCaptionStyle{ruled}{labelfont=normalfont,labelsep=colon,strut=off} 
\floatstyle{ruled}
\newfloat{listing}{tb}{lst}{}
\floatname{listing}{Listing}
\title{MGFN : Magnitude-Contrastive Glance-and-Focus Network \\for Weakly-Supervised Video
Anomaly Detection}
\author{
    Yingxian Chen\textsuperscript{1},
    Zhengzhe Liu\textsuperscript{2},
    Baoheng Zhang\textsuperscript{1},
    Wilton Fok\textsuperscript{1},
    Xiaojuan Qi\textsuperscript{1},
    Yik-Chung Wu\textsuperscript{1}
}
\affiliations{
    \textsuperscript{\rm 1}Department of Electrical and Electronic Engineering, The University of Hong Kong\\
    \textsuperscript{\rm 2}The Chinese University of Hong Kong\\

    \{carolcyx, cheungbh\}@hku.hk, \{wtfok, xjqi, ycwu\}@eee.hku.hk, zzliu@cse.cuhk.edu.hk

}

\usepackage{bibentry}

\begin{document}

\maketitle

\begin{abstract}
Weakly supervised detection of anomalies in surveillance videos is a challenging task. Going beyond existing works that have deficient capabilities to localize anomalies in long videos, we propose a novel glance and focus network to effectively integrate spatial-temporal information for accurate anomaly detection. In addition, we empirically found that existing approaches that use feature magnitudes to represent the degree of anomalies typically ignore the effects of scene variations, and hence result in sub-optimal performance due to the inconsistency of feature magnitudes across scenes. To address this issue, we propose the Feature Amplification Mechanism and a Magnitude Contrastive Loss to enhance the discriminativeness of feature magnitudes for detecting anomalies. Experimental results on two large-scale benchmarks UCF-Crime and XD-Violence manifest that our method outperforms state-of-the-art approaches. 
\end{abstract}

\section{Introduction}

\noindent Detection of anomalies in surveillance videos is an important research topic \cite{survey1,survey2}. The abnormal events, such as accidents, fighting, arson, and banditry, may cause social harm and even threaten human life \cite{survey3}. 

\begin{figure}[!t]
\centering
\includegraphics[width = 8.5 cm]{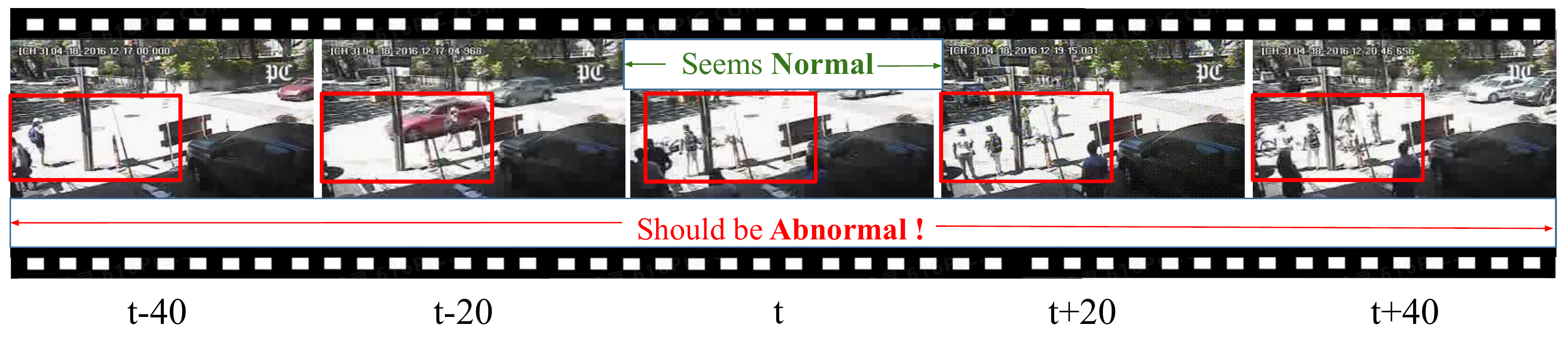}
\vspace{-6mm}
\caption{Long-term temporal context information plays a significant role in surveillance video-based anomaly detection. It seems nothing happens in frame $t$, but the opposite conclusion can be easily derived considering other frames. }
\vspace{-3mm}
\label{fig:suvideo}
\end{figure}

Unfortunately, it is extremely challenging to identify and locate anomalies in a long video \cite{challen-1,challen-2,challen-3,challen-4}. Firstly, ``anomaly'' is a relative term defined against ``normality''. Hence, as shown in Fig.~\ref{fig:suvideo}, it is unreasonable to predict the anomaly based upon only a single or a few nearby frames without the necessary knowledge of the ``normality''. In addition, ``anomalies'' include a variety of events mentioned above, so it is challenging to construct a unified representation for all the varying events.  
Besides, it is tedious and laborious to prepare for the frame-level annotation. Therefore, network training is more feasible under  video-level weak supervisions. 

To mitigate the above challenges, existing methods can be roughly cast into two branches. One stream learns to detect anomalies by constructing spatial-temporal architectures ~\cite{MIST,motion-aware,RTFM,Not-only-look,LCTR}. Although such models work well for short videos, they typically struggle in tackling long videos where  anomalous frames occupy only a small portion, due to the lack of global context awareness and specific focus on the abnormal frames. 

\begin{figure*}[!t]
\centering
\includegraphics[width = 16 cm]{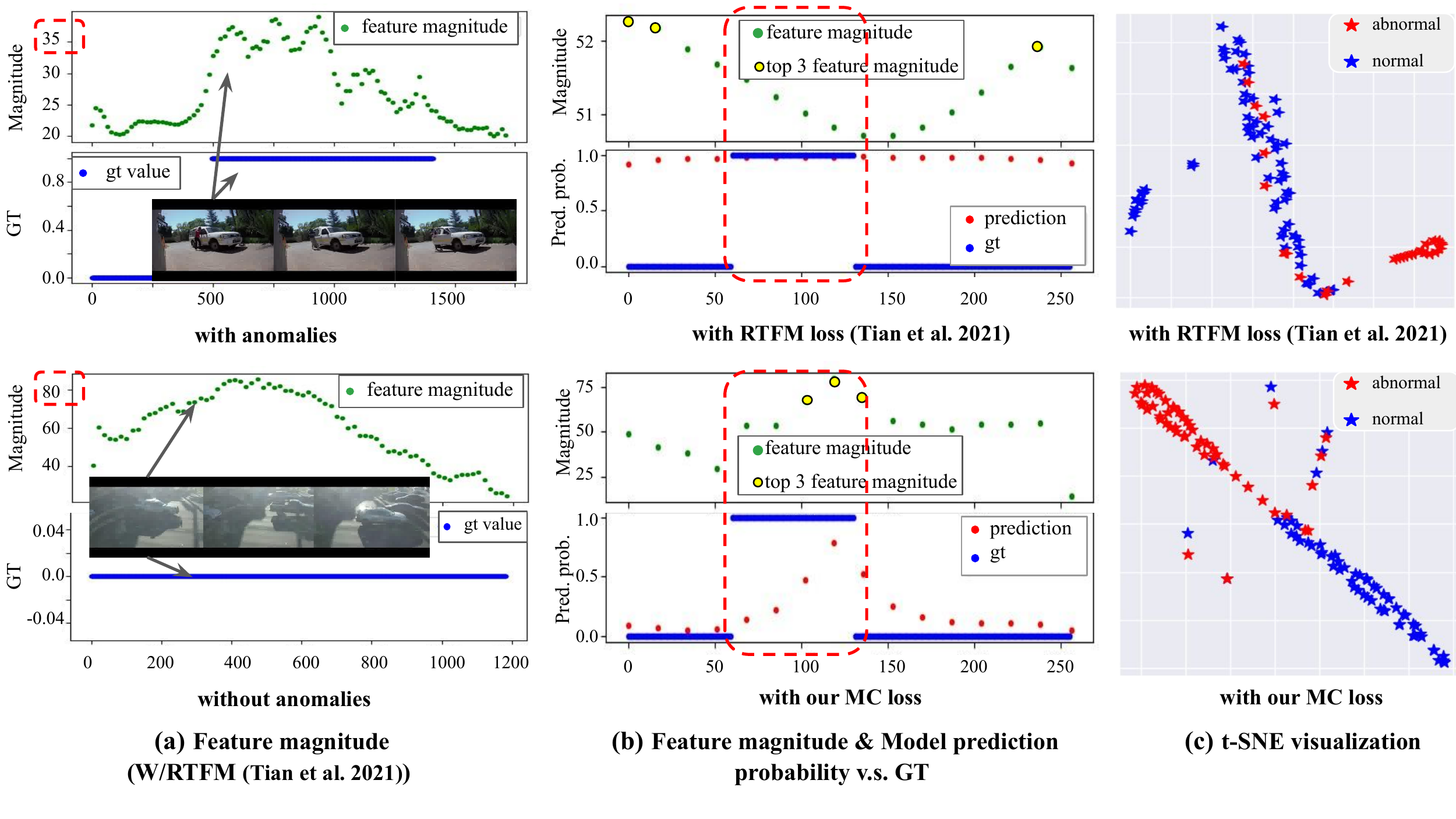}
\vspace*{-2mm}
\caption{(a) With RTFM loss \cite{RTFM}, normal feature magnitudes can be larger than abnormal ones due to the various scene attributes. (b) Our MC loss can generate more consistent feature magnitudes with anomalies. (c) Based on 100 random samples, our MC loss achieves the better separability of normal and abnormal features. }
\vspace*{-3mm}
\label{fig:intro_visulization}
\end{figure*}

The second branch learns to distinguish anomaly from normal by designing loss functions \cite{RTFM,cleaning-label,MIST,claws}. 
For example, the most recent work \cite{RTFM} proposes a Robust Temporal Feature Magnitude (RTFM) loss to push abnormal feature magnitudes to be larger and normal ones in the opposite direction. 
It might be true that within the same video sequence or in similar scenes, abnormal features attain larger magnitudes than normal ones. 
However, we empirically found that besides the anomaly, the feature magnitude also depends on other attributes of the video such as the object movements, number of objects and persons in the scene, etc. As shown in Fig.~\ref{fig:intro_visulization} (a), the feature magnitudes in the normal video (the below) with substantial object movement are even larger than the abnormal one (the above). 

Therefore, it is unreasonable to simply encourage the abnormal features to be larger and normal features to be smaller using the RTFM loss \cite{RTFM}, since this target is not consistent with the inherent inter-video magnitude distribution and thus harms network training. 
We further found that even in the same video sequence, as shown in Fig.~\ref{fig:intro_visulization} (b), some normal features learned in \cite{RTFM} (green points outside the red bounding box) attain similar or even larger magnitudes compared with abnormal ones (green points in the red bounding box). In addition, the unsatisfying feature separability (Fig.~\ref{fig:intro_visulization} (c)) indicates that RTFM \cite{RTFM} cannot effectively separate normal and abnormal features. 

To address the aforementioned issues, we propose a Magnitude-Contrastive Glance-and-Focus Network (MGFN) for anomaly detection. Inspired by global-to-local information integration mechanism in the human vision system for detecting anomalies in a long video, MGFN first glances the whole video sequence to capture long-term context information, and then further addresses each specific portion for anomaly detection. Different from existing works that simply fuse the spatial-temporal features, our strategy helps the network to first get an overview of the scene, and then detect scene-adaptive anomaly based on global knowledge as a prior. To the best of our knowledge, this is the first work to explore the glance and focus mechanism for video anomaly detection. 

In addition, we propose a simple yet effective Feature Amplification Mechanism (FAM) to enhance the discriminativeness of feature magnitude for anomaly detection. 
More importantly, unlike the RTFM loss \cite{RTFM} that simply pushes normal and abnormal features to the opposite directions without considering different scene attributes, we propose a Magnitude Contrastive (MC) loss to learn a scene-adaptive cross-video magnitude distribution. It encourages the similarity of the feature magnitudes for videos in the same category and the separability of those between normal and abnormal videos. Our MC loss improves the consistency between the feature magnitude and anomaly 
(Fig.~\ref{fig:intro_visulization}(b)), and hence benefits the separability of normal and abnormal features (Fig.~\ref{fig:intro_visulization}(c)). 

To summarize, our contributions include:
\begin{itemize}
\item To the best of our knowledge, we first propose a Magnitude-Contrastive Glance-and-Focus Network (MGFN) to mimic the global-to-local information integration mechanism of the human visual system which first glance the video and then focus on the local portion for video anomaly detection.
\item We design the Feature Amplification Mechanism (FAM) to enhance the feature learning and a Magnitude Contrastive Loss to encourage the separability between normal and abnormal features. 
\item Our proposed MGFN achieves 86.98\% (AUC) and 80.11\% (AP) on two large-scale datasets UCF-crime \cite{RWAD} and XD-violence \cite{Not-only-look} respectively, outperforming the state of the arts by a large margin. The codes are available in \url{https://github.com/carolchenyx/MGFN.git}.
\end{itemize}

\section{Related Work}

\subsection{Video Anomaly Detection (tasks)}
\noindent Due to the huge amount of training data in surveillance video, it is very labor-intensive and time-consuming to annotate every frame of the video. Therefore, researchers tend to focus on the one-class \cite{one-class} or unsupervised setting to learn anomaly detection without any annotation, and the weakly supervised setting where only the video-level annotations are available. 

In the one-class learning branch, Luo \etal designs a ConvLSTM network to learn the normal segments \cite{un-rhc}. Zhao \etal, Ionescu \etal and Chang \etal adopt Auto-Encoder (AE) networks to reconstruct features of normal frames \cite{un-sta,un-oada,un-cdda}. Other works utilize the memory mechanism to memorize normal patterns \cite{un-mem1,un-mem2,un-mem3,un-mem4} and use meta-learning \cite{un-fsad} to enhance model's generalization capability to the unseen normal scenarios.  

In the weakly supervised learning branch, the early work \cite{RWAD} uses a multiple instance learning method to localize anomalous clips in videos. Recently, Zhong \etal adopts graph convolution networks to learn abnormal events \cite{GCLNC}. However, the generalization capability of such a model is far from satisfactory. To address this issue, Ramachandra \etal builds Siamese network to learn the normal feature simulation  builds Siamese network. Afterwards, Wan \etal and Zaheer \etal propose the clustering-based frameworks to distinguish anomalous events \cite{weakly-wsvad,claws}. Many recent works propose some network architectures to learn the spatial-temporal feature ensemble \cite{wsstad,MIST,RTFM,LCTR,MSL,IBL}, \eg, Tian \etal~\cite{RTFM} exploits an Robust Temporal Feature Magnitude (RTFM) to leverage the feature magnitude for anomaly detection. 
To better learn the temporal relation and feature discrimination, Wu \etal adds the information of the previous few snippets \cite{LCTR} and achieves the state-of-the-art performance on UCF-Crime dataset.

In this work, we focus on the weakly-supervised anomaly detection due to its good balance between annotation burden and detection performance.

\subsection{Vision Transformer (approaches)}
Transformers are first utilized in natural language processing (NLP) such as machine translation \cite{trans1} and plaintext compression \cite{trans2}. They have achieved great improvements on NLP-related tasks and have been proved that they have powerful feature representation capacity \cite{trans-survey}.

Inspired by the achievements in NLP, transformer architectures have recently been exploited in computer vision tasks including image classification \cite{ffn}, image processing \cite{vt-1}, object detection \cite{vt-2}, semantic segmentation \cite{vt-3}, action classification \cite{vt-4}, and video processing \cite{vt-5}.

Compared with images, video data has one additional temporal dimension. Recently, many researchers have explored transformer-based architectures to learn the spatial and temporal dependencies for video processing tasks. Fayyaz \etal and Chi \etal use transformer to improve video recognition performance \cite{trans-vi1,trans-vi2}. Yin \etal builds a spatial-temporal transformer to capture the spatial and temporal information for video object detection \cite{trans-vi3}. As for untrimmed videos, Seong \etal proposes the multi-task transformer network to reduce the feature redundancy and learn the relationship between different dimensions \cite{trans-vi4}. Inspired by the capabilities of a transformer, in this paper, we propose transformer-based blocks to glance at the whole video globally and then steer attention to each video portion, imitating human beings' global-to-local vision system for anomaly detection.

\section{Our Approach}

\begin{figure*}[!t]
\centering
\includegraphics[width=16cm]{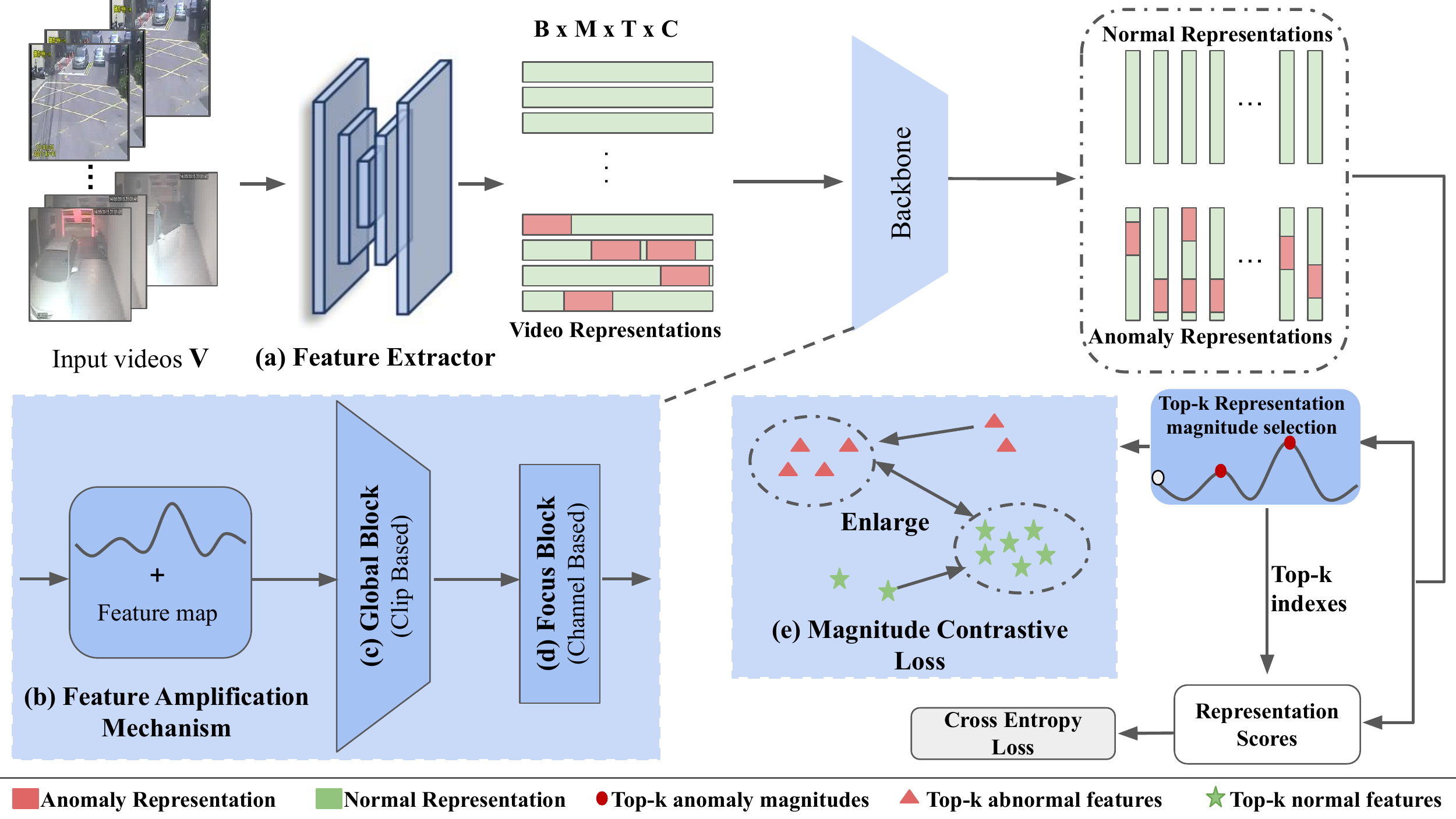}
\vspace*{-2mm}
\caption{The overview of our MGFN network architecture. The framework takes {\small$B/2$} normal videos and {\small$B/2$} abnormal videos as input. After (a) Feature extractor, (b) Feature Amplification Mechanism (FAM) calculates the feature magnitude and incorporates it as a residue explicitly. Then (c) Glance Block (GB) and (d) Focus Block (FB) extract the global context information and enhance the local feature respectively. (e) Magnitude Contrastive (MC) loss encourages the separability of normal and abnormal features by shrinking the intra-category feature magnitude distances and enlarge the inter-category differences using the top-k normal and abnormal feature magnitudes.}
\vspace*{-3mm}
\label{fig:model-structure}
\end{figure*}

\begin{figure}
\centering
\includegraphics[width=8cm]{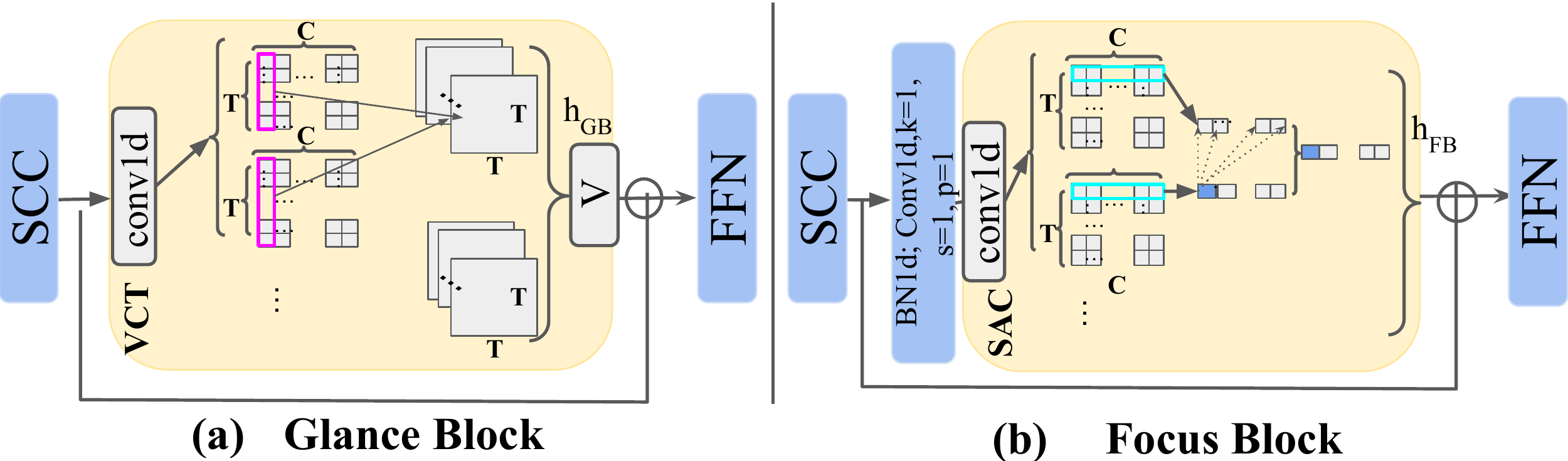}
\vspace{-2mm}
\caption{Architectures of Glance Block and Focus Block.}
\vspace{-4mm}
\label{fig:block-structure}
\end{figure}

\subsection{Overview}
The architecture of our framework is shown in Fig.~\ref{fig:model-structure}. First, feature extractor takes {\small $B$} untrimmed videos {\small$V$} with video-level annotation as input (see Fig.~\ref{fig:model-structure} (a)), where {\small$V_i\in\mathbb{R}^{N_i \times H\times W\times 3}$} and {\small$N_{i},H,W$} denote the number of frames, height and width of {\small$V_{i}$} respectively. Then we evenly segment each video sequence into {\small$T$} clips and denote the feature map from feature extractor as {\small$F=\{f^{i,t}$}, where {\small$i\in[1,B], t\in [1,T] \}\in\mathbb{R}^{B \times T\times P\times C}$}, where {\small$P$} is the number of crops in each video clip \cite{RWAD}, and {\small$C$} is the feature dimension. {\small$f^{i,t}\in\mathbb{R}^{P\times C}$} denotes the feature of the {\small$t^{th}$} video clip in {\small$V_{i}$}.

Taking feature map {\small$F$} as input, Feature Amplification Mechanism (FAM) (Fig.~\ref{fig:model-structure} (b)) explicitly calculates the feature norm {\small$M$} to enhance {\small$F$}. Then Glance Block (GB) and Focus Block (FB) (Fig.~\ref{fig:model-structure} (c,d)) integrate the global and local features built upon video clip-level transformer (VCT) and self-attentional convolution (SAC) respectively. Instead of simply maximizing the magnitude of abnormal features {\small$f_a$} and minimizing normal ones {\small$f_n$} like that in \cite{RTFM}, we design a Magnitude Contrastive (MC) loss (Fig.~\ref{fig:model-structure} (e)) to maximize the separability of normal features magnitudes and abnormal ones. In the following, we will detail each component of MGFN and the loss functions in the network training.

\subsection{Feature Amplification Mechanism (FAM)} 
As shown in Figure.~\ref{fig:model-structure} (b), FAM first explicitly calculates the feature norm {\small$M^{i,t}$} of {\small$f^{i,t}$} as Equation (\ref{MFM-1}):
{\small\begin{equation}
\begin{aligned}
    M^{i,t} =  \biggl(\sum_{c=1}^C \left|f^{i,t,c}\right|^2\biggl)^\frac{1}{2} \in \mathbb{R}^{1 \times 1 \times P \times 1}
\label{MFM-1}
\end{aligned}
\end{equation}}
where $c$ denotes the feature dimension index. 

Afterwards, FAM derives the enhanced features  {\small$F_{FAM}=\{f^{i,t}_{FAM}\}$} by adding the 1D-Convolution modulated feature norm, Conv1D({\small$M^{i,t}$}) to {\small$f^{i,t}$} as a residue as shown in Equation (\ref{MFM-2}): 
{\small\begin{equation}
\begin{aligned}
     f^{i,t}_{FAM} =  f^{i,t}+ \alpha \text{Conv1D}(M^{i,t}) \in \mathbb{R}^{1 \times 1 \times P \times C}
\label{MFM-2}
\end{aligned}
\end{equation}}

\noindent where $\alpha$ is a hyper-parameter to control the effect of the norm term, and Conv1D is a single-layer 1D convolutional network to modulate the feature norm for each dimension. 

Without affecting the feature map dimension, FAM amplifies the feature map by explicitly incorporating the feature norm, which is a unified abnormal representation, into the network to benefit the Magnitude Contrastive Loss that will be discussed later. 

\subsection{Glance Block}
\label{glance}
The architecture of the Glance Block is shown in Fig.~\ref{fig:block-structure} (a). To reduce the computational burden, we first use a convolution to decrease the feature map dimension from {\small$C$} in {\small$F_{FAM}$} to {\small$C/32$}. After a short-cut convolution (SCC) that outputs a feature map {\small$F_{scc\_GB}\in\mathbb{R}^{B \times T \times P \times C/32}$}, we construct a video clip-level transformer (VCT) to learn the global correlation among clips. 
Specifically, we establish an attention map {\small$A\in \mathbb{R}^{1 \times T \times T\times P}$} to explicitly correlate the different temporal clips. 
{\small\begin{equation}
\begin{aligned}
    A^{i,t_1,t_2}=\sum_{c=1}^C Q(F_{scc\_GB}^{i,t_1,c})K(F_{scc\_GB}^{i,t_2,c}) 
\label{GB-2}
\end{aligned}
\end{equation}}
where {\small$t_1,t_2\in [1,T]$}, {\small$Q,K$} are 1D ``query'' and ``key'' convolutions of transformer.  

Next, we use the softmax normalization to generate {\small$a\in \mathbb{R}^{1 \times T \times T\times P}$} where {\small$a^{i,t_1,:}$} represents how other clips associate with clip $t_1$. 
{\small\begin{equation}
\begin{aligned}
    a^{i,t_1,t_2}=\frac{e^{A^{i,t_1,t_2}}}{\sum_{t_2=1}^T e^{A^{i,t_1,t_2}}} 
\label{GB-3}
\end{aligned}
\end{equation}}
The output of VCT {\small$F_{att\_{GB}}\in \mathbb{R}^{1 \times T \times P \times C/32}$} is the weighted average of all clips in the long video containing both normal and abnormal (if exists) ones: 
{\small\begin{equation}
\begin{aligned}
    F_{att\_{GB}}^{i,t_1,c}=\sum_{t_2=1}^T a^{i,t_1,t_2} V(F_{scc\_GB}^{i,t_2,c}) 
\label{GB-1}
\end{aligned}
\end{equation}}
where {\small$V$} is the 1D ``value'' convolution of transformer. 
Therefore, Glance Block provides the network with the knowledge of ``what the normal cases are like'' to better detect the abnormal events. In addition, it helps the network to better utilize the long-term temporal contexts. 

Glance Block contains an additional Feed-Forward Network (FFN) including two fully-connected layers and a GeLU non-linear function to further improve the model's representation capability. The output feature map {\small$F_{GB}$} is fed to the following Focus Block. 

\subsection{Focus Block}
\label{focus}
As shown in Fig.~\ref{fig:block-structure} (b), the Focus Block (FB) consists of a short-cut convolution (SCC), a self-attentional convolution (SAC), and a Feed-Forward-Network (FFN). With {\small$F_{GB}$} as input, we first increase the channel number to {\small$C/16$} with a convolution. Then SSC generates the feature map {\small$F_{scc\_FB}$}. 

Inspired by the self-attention mechanism, we propose a self-attentional convolution (SAC) to enhance the feature learning in each video clip. Specifically, we exploit {\small$F_{scc\_FB}$} as both the feature map and convolution kernel, and formulate this step to be a convolution with kernel$\_$size=5 as Equation (\ref{FB-1}) and (\ref{FB-2}). 
{\small\begin{equation}
\begin{aligned}
    F_{sac\_FB}=F_{scc\_FB} \circledast F_{scc\_FB}  \in \mathbb{R}^{B \times T \times P\times C/16}
\label{FB-1}
\end{aligned}
\end{equation}}
where 
{\small\begin{equation}
\begin{aligned}
    F_{sac\_FB}^{i,t,k_1}=\sum_{k_1,k_2 = 0}^{C/16} F_{scc\_FB}^{i,t,k_1}  F_{scc\_FB}^{i,t,k_2} \in \mathbb{R}^{1 \times 1 \times P\times 1}
\label{FB-2}
\end{aligned}
\end{equation}}

Like self-attention, our self-attentional convolution allows each channel to get access to the nearby channels to learn the channel-wise correlation without any learnable weight. After a two-layered FFN, FB outputs the feature map {\small$F_{FB}$}.

\subsection{Loss Functions}

In this section, we introduce our loss functions. Since anomaly detection is a binary-class classification problem, a natural loss function is the sigmoid cross-entropy loss for the predicted score like existing works \cite{weakly-wsvad,GCLNC,Not-only-look,RTFM}: {\small$ L_{sce} = -y\log(s^{i,j})-(1-y)\log(1-s^{i,j})$}, where $y$ is the video-level ground truth ({\small$y=1$} indicates abnormal) and {\small$s^{i,j}$} is the predicted abnormal probabilities of clip {$j$}.  

\subsubsection{Magnitude Contrastive Loss}

To better encourage the feature separability, we propose a Magnitude Contrastive (MC) Loss as Equation (\ref{MCL}). 
{\small\begin{equation}
\begin{aligned}
     L_{mc} = &\sum_{p,q=0}^{B/2}(1-l)(D(M^p_{n},M^q_{n})) + \sum_{u,v=B/2}^{B}(1-l)(D\\
     &(M^u_{a},M^v_{a})) + \sum_{p=0}^{B/2}\sum_{u=B/2}^{B}l(\text{Margin} - D(M_{n}^p,M_{a}^u))
\label{MCL}
\end{aligned}
\end{equation}}
Note that we sample {\small$B/2$} normal videos and {\small$B/2$} abnormal videos in a training batch {\small$B$}. $p,q$ are the indexes of normal clips, and $u,v$ are for abnormal clips. {\small$M_a$} indicates the top-k feature magnitudes of abnormal clips and {\small$M_n$} means those of normal clips.  {\small$D(\cdot,\cdot)$} is a distance metric that will be described below. 
$l$ is an indicator function where $l=1$ denotes a pair of normal and abnormal clips $p,u$ is sampled. In this case, {\small$L_{mc}$} increases the feature magnitude distance of them. {\small$l=0$} denotes that the two sampled clips $p,q$ or $u,v$ are both normal or abnormal, where {\small$L_{mc}$} groups them together. 

Instead of simply increasing the feature magnitudes of abnormal clips and reducing those of normal clips like Tian \etal \cite{RTFM}, our MC loss learns a scene-adaptive cross-video feature magnitude distribution. Specifically, MC loss does not force all abnormal features in different scenes to be larger than normal ones. Instead, it encourages the model to separate them with the proper distribution. For example, we allow a normal video with substantial movement to have larger feature magnitudes than an abnormal one. Accompanying with other features learned by the network, such as the scene- and movement-related features, the model can still correctly predict the anomaly. In addition, we further found that in the same video and similar scenes, the abnormal feature magnitudes are typically larger than normal ones (see Fig.~\ref{fig:intro_visulization} (c)), which is consistent with the aim of RTFM loss \cite{RTFM}]. 

{\small$D(M^p_n,M^q_n)$} is defined in Equation (\ref{MCL_D}), and {\small$D(M^u_a,M^v_a)$} is defined similarly. 
{\small\begin{equation}
\begin{aligned}
D(M^p_n,M^q_n) = \min_{t={0,\dots,T},r={0,\dots,T}}   (\vmathbb{1} (\left\Vert f^{p,t}_{FB}\right\Vert_2) -  \vmathbb{1} (\left\Vert f^{q,r}_{FB}\right\Vert_2)) 
\label{MCL_D}
\end{aligned}
\end{equation}}
{\small$\vmathbb{1}$} is a top-k mean function where {\small$\vmathbb{1}(\left\Vert f^{p,t}_{FB}\right\Vert_2)$} is one of the top-k feature magnitudes among {\small$f^{p,\{1,\dots,T\}}_{FB}$}. {\small$D(\cdot,\cdot)$} derives the feature distance based on the top-k-largest-feature-magnitude clips throughout the {\small$T$} clips of each video. Choosing top-k clips instead of the whole video sequence benefits the model training under the video-level weak supervision. Due to the absence of the clip-level abnormality ground truth, top-k selection helps {\small$L_{mc}$} to focus on the clips that are most likely to be abnormal in the abnormal videos, as well as the hardest cases in normal videos. At the same time, we also select the maximum-distance pair among the top-k normal and top-k abnormal clips and encourage their similarity using {\small$L_{mc}$}.

Similarly, the distance {\small$D(M^p,M^u)$} of normal feature $p$ and abnormal $u$ is defined as Equation (\ref{MCL_D2}).
{\small\begin{equation}
\begin{aligned}
D(M^p_n,M^u_a) = \max_{t={0,\dots,T},r={0,\dots,T}}   (\vmathbb{1} (\left\Vert f^{p,t}_{FB}\right\Vert_2) -  \vmathbb{1} (\left\Vert f^{u,r}_{FB}\right\Vert_2)) 
\label{MCL_D2}
\end{aligned}
\end{equation}}

\subsubsection{Overall Loss Functions} Following Sultani \etal \cite{RWAD}, we adopt temporal smoothness loss {\small$L_{ts} = \sum^{T}_{j=1} s^{i,j}_a $}, and sparsity loss {\small$L_{sp} = \sum^{T}_{j=1} (s^{i,j}_a-s^{i,j+1}_a)^2$}, where {\small$s_a$} denotes the prediction score of abnormal clips. They work as regularizations to smooth the predicted scores of adjacent video clips. Thus, the total loss in the model training is:
{\small$ L = L_{sce} + \lambda_1 L_{ts} + \lambda_2 L_{sp} + \lambda_3 L_{mc}$}, where $\lambda_1,\lambda_2,\lambda_3$ are loss weights to balance the loss terms.

\section{Experiment}
\subsection{Benchmarks and Evaluation Metrics}
We consider two benchmarks in our analysis, UCF-Crime \cite{RWAD} and XD-Violence \cite{Not-only-look}. 
For these two benchmarks, only video-level annotations are provided. The abnormal videos contain both normal and abnormal frames, and the normal videos only contain normal frames. 
Following \cite{claws,Not-only-look,RTFM,LCTR,motion-aware}, we adopt Area Under the Curve (AUC) under the Receiver Operating Characteristic (ROC) curve as an evaluation metric for the UCF-Crime, and utilize Average Precision (AP) as the metirc for the XD-Violence. Larger AUC and AP indicate the better performance of the model. 

\subsection{Implementation Details}
Our proposed method is implemented in PyTorch \cite{pytor}. The feature extractors are I3D \cite{i3d} and VideoSwin \cite{videoswin}. The hyper-parameters are set as {\small$T=32$}, {\small$P=10$}, {\small$\alpha=0.1$}, {\small$k=3$}, {\small$\lambda_1=\lambda_2=1$}, {\small$\lambda_3=0.001$}. To train the network, we used Adam optimiser \cite{adam} with a weight decay of 0.0005 and a learning rate of 0.001. The batch size {\small$B$} in the training is 16 and each batch consists of randomly selected 8 normal videos and 8 abnormal ones.

\subsection{Results on UCF-Crime}
As shown in Table \ref{table:ucf-results}, our result outperforms all the existing one-class baselines \cite{un1,SCR,un2}, unsupervised work \cite{one-class} and weakly-supervised works \cite{RWAD,MIST,Not-only-look,RTFM,one-class} by a large margin. Specifically, our approach achieves better performance consistently compared with \cite{Not-only-look,RTFM,LCTR} that utilize spatial-temporal feature learning, indicating the effectiveness of our Glance and Focus Blocks. 
Besides, with I3D features, our method outperforms RTFM \cite{RTFM} by 2.85\% which utilizes the RTFM loss, manifesting the superiority of our proposed MC loss. With VideoSwin features, our approach even surpasses the SOTA approach MSL \cite{MSL} by 1.05\% AUC, which is already a significant improvement in the weakly-supervised video anomaly detection field. 

\vspace*{-1mm}
\subsection{Results on XD-Violence}
Table \ref{table:xd-results} shows the results on XD-Violence dataset. Again, we achieve the superior performance over all the existing works. Specifically, our approach outperforms the spatial-temporal feature ensemble approaches \cite{Not-only-look,LCTR,RTFM} consistently, demonstrating the high performance of our Glance-Focus mechanism. Thanks to our MC loss, we outperform SOTA approaches RTFM \cite{RTFM} by 1.38\% with I3D features and MSL \cite{MSL} by more than 1.53\% AP with VideoSwin features, indicating the effectiveness of MC loss over their RTFM loss. The consistent superiority of our approach demonstrates the effectiveness of our proposed model for weakly-supervised video anomaly detection.

\setlength{\tabcolsep}{1pt}
\begin{table}[!t]{\small
\begin{center}
\begin{tabular}{ccccc}
\hline\noalign{\smallskip}
Supervision & Method & Features & T=32 & AUC(\%) \\
\noalign{\smallskip}
\hline
\noalign{\smallskip}
{} & SVM Baseline & - & - & 50.00 \\
{} & SSV (2018) & - & - & 58.50\\
One-class  & BODS (2019) & I3D-RGB & - & 68.26\\
{} & GODS (2019)  & I3D-RGB & - & 70.46\\
{} & SACR (2020)  & - & - & 72.70\\
{} & Zaheer et al. (2022) & ResNext & - & 74.20\\
\hline
\makecell[c]{Un-\\supervised} & \makecell[c]{Zaheer et al. (2022)} & ResNext & - & 71.04\\
\hline
{} & Sultani \etal (2018) & C3D-RGB & \checkmark & 75.41\\
{} & Sultani \etal (2018) & I3D-RGB & \checkmark & 77.92\\
{} & IBL \etal (2019) & C3D-RGB & \checkmark & 78.66\\
{} & Zaheer et al. (2022) & ResNext & - & 79.84\\
{} & GCN (2021) & TSN-RGB & - & 82.12\\
{} & MIST (2021) & I3D-RGB & - & 82.30\\
Weakly- & Wu \etal (2020) & I3D-RGB & \checkmark & 82.44\\
supervised & CLAWS (2021) & C3D-RGB & - & 82.30\\
{} & RTFM (2021) $*$ & VideoSwin-RGB & \checkmark & 83.31\\
{} & RTFM (2021) & I3D-RGB & \checkmark & 84.03\\
{} & Wu and Liu (2021) & I3D-RGB & \checkmark & 84.89\\
{} & MSL (2022) & I3D-RGB & \checkmark &  85.30\\
{} & MSL (2022) & VideoSwin-RGB & \checkmark &  85.62\\
{} & \textbf{MGFN(Ours)} & \textbf{I3D-RGB} & \checkmark & \textbf{86.98}\\
{} & \textbf{MGFN(Ours)} & \textbf{VideoSwin-RGB} & \checkmark & \textbf{86.67}\\
\hline
\end{tabular}
\caption{Comparison with existing works on UCF-Crime dataset. Our results outperform all prior works with different features, indicating that our methods' effectiveness. (T=32 means a video is divided into 32 non-overlap clips. $*$ means the result is reported by \cite{MSL})}
\vspace*{-4mm}
\label{table:ucf-results}
\end{center}}
\end{table}
\setlength{\tabcolsep}{1pt}

\setlength{\tabcolsep}{1pt}
\begin{table}[!t]{\small
\begin{center}
\begin{tabular}{ccccc}
\hline\noalign{\smallskip}
Supervision & Method & Features & T=32 &AP(\%) \\
\noalign{\smallskip}
\hline
\noalign{\smallskip}
{} & Wu \etal (2020) & C3D-RGB & $-$ & 67.19 \\
{} & Sultani \etal (2020) $\diamond$ & I3D-RGB & $-$ & 73.20 \\
{}  & MSL (2022) & C3D-RGB & $-$  &  75.53\\
{} & Wu \etal (2020) $*$ & I3D-RGB & \checkmark  &75.68 \\
Weakly- & Wu and Liu (2021) & I3D-RGB & \checkmark & 75.90\\
supervised & RTFM (2021) & I3D-RGB & \checkmark & 77.81\\
{} & MSL (2022) & I3D-RGB & \checkmark &78.28\\
{} & MSL (2022) & VideoSwin-RGB & \checkmark &78.58\\
{} & \textbf{MGFN(Ours)} & \textbf{I3D-RGB} & \checkmark  & \textbf{79.19}\\
{} & \textbf{MGFN(Ours)} & \textbf{VideoSwin-RGB} & \checkmark & \textbf{80.11}\\
\hline
\end{tabular}
\caption{Frame-level AP on XD-Violence dataset. $*$ means the result is derived using I3D-RGB features from \cite{RTFM}. $\diamond$ means result was reported by \cite{Not-only-look}.}
\vspace{-3mm}
\label{table:xd-results}
\end{center}}
\end{table}
\setlength{\tabcolsep}{1pt}

\subsection{Ablation Studies}
We conduct extensive ablation studies to validate the effectiveness of the key components in our MGFN: Glance-Focus mechanism, FAM, and MC loss. 

\subsubsection{Glance-Focus Mechanism}
\begin{figure}[!t]
\centering
\includegraphics[width=8.5cm]{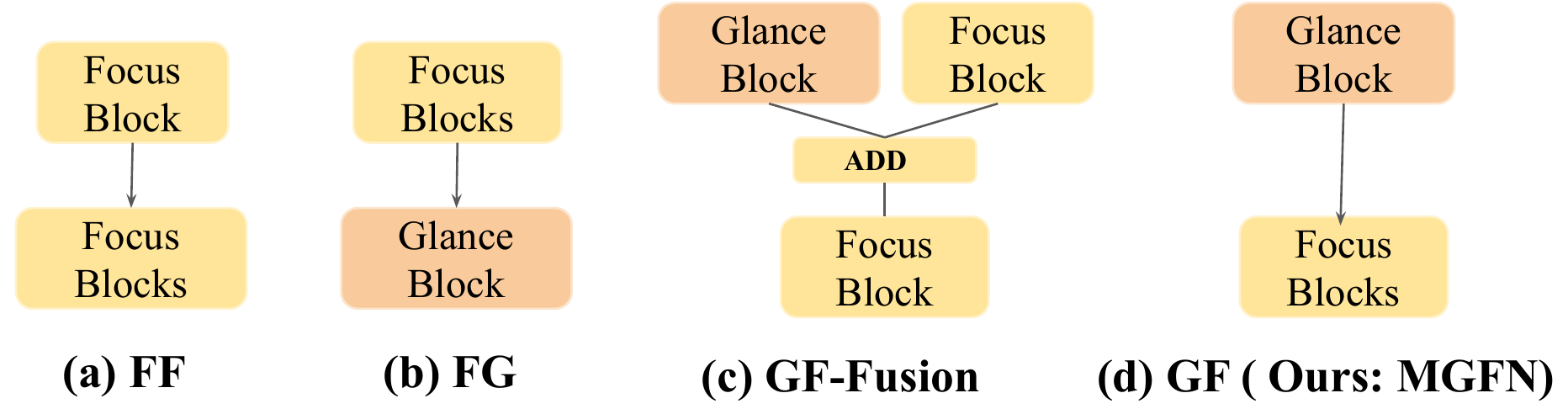}
\vspace{-5mm}
\caption{\textbf{Glance-Focus structure (GF) analysis.} (G: Glance Block, F: Focus Block) (a)-(d): Different designed network structures.}
\vspace{-3mm}
\label{fig:blocks-structure}
\end{figure}

To verify the effectiveness of the Glance-Focus mechanism to first glance the whole video and then focus on local video portions, we construct four baselines including three different organizations of our Glance Block (GB) and Focus Block (FB), and the state-of-the-art work \cite{RTFM}. As shown in Fig.~\ref{fig:blocks-structure} (a to d), 
FF means two cascaded FBs, FG stands for the cascaded F and G, and GF-Fusion means G and F are summed together. Finally, GF means our Glance-Focus mechanism where G comes first and then F follows.

\begin{figure}[!t]
\centering
\includegraphics[width=8.5cm]{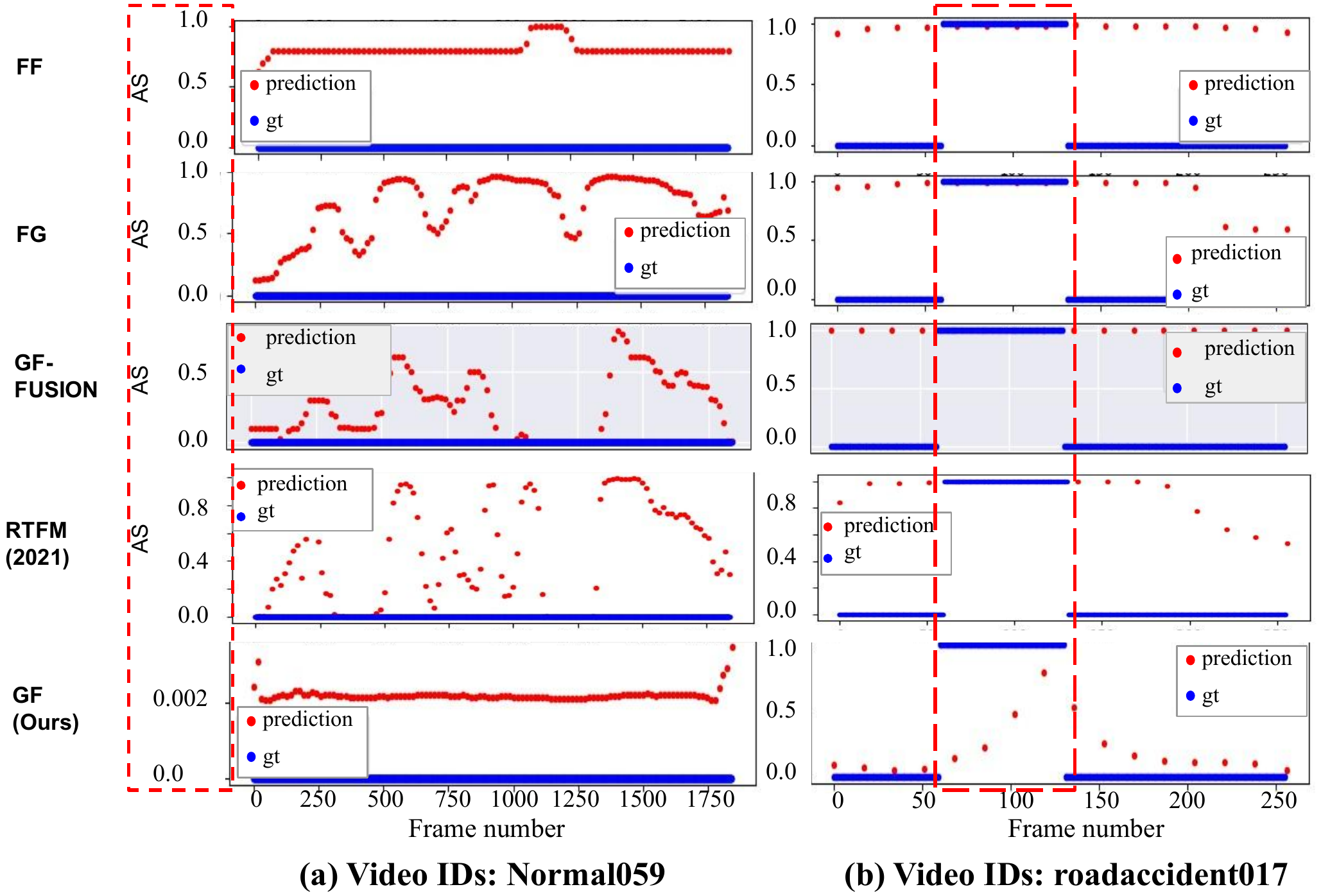}
\vspace{-5mm}
\caption{\textbf{Glance-Focus structure (GF) analysis.} (AS: anomaly score) (a) and (b) are the predicted anomaly scores (red points) of different designed structures (FF, FG, GF-FUSION and GF) and baseline (RTFM) v.s. ground truth (blue lines) on the UCF-crime dataset (``normal059'' and ``roadaccident017''). The visualizations show that \textbf{Glance-Focus (GF)} is able to distinguish the normal and anomalies effectively. }
\vspace{-1mm}
\label{fig:ablation-1}
\end{figure}

As shown in Fig.~\ref{fig:ablation-1}, FF, FG, GF-Fusion and the state-of-the-art work \cite{RTFM} cannot make the consistent prediction throughout a normal video (see Fig.~\ref{fig:ablation-1} (a)). In addition, FF, FG and GF-Fusion fail to spot out the abnormal frames in Fig.~\ref{fig:ablation-1} (b), and \cite{RTFM} creates some false-positive predictions (see the points just outside the red bounding box in Fig.~\ref{fig:ablation-1} (b) RTFM \cite{RTFM}). On the contrary, our MGFN with the Glance-Focus mechanism can not only produce the stable predictions throughout a normal video, but also successfully detect the abnormal events in the abnormal video (see Fig.~\ref{fig:ablation-1} (a,b) GF (Ours)).

\begin{figure}[!t]
\centering
\includegraphics[width=8.5cm]{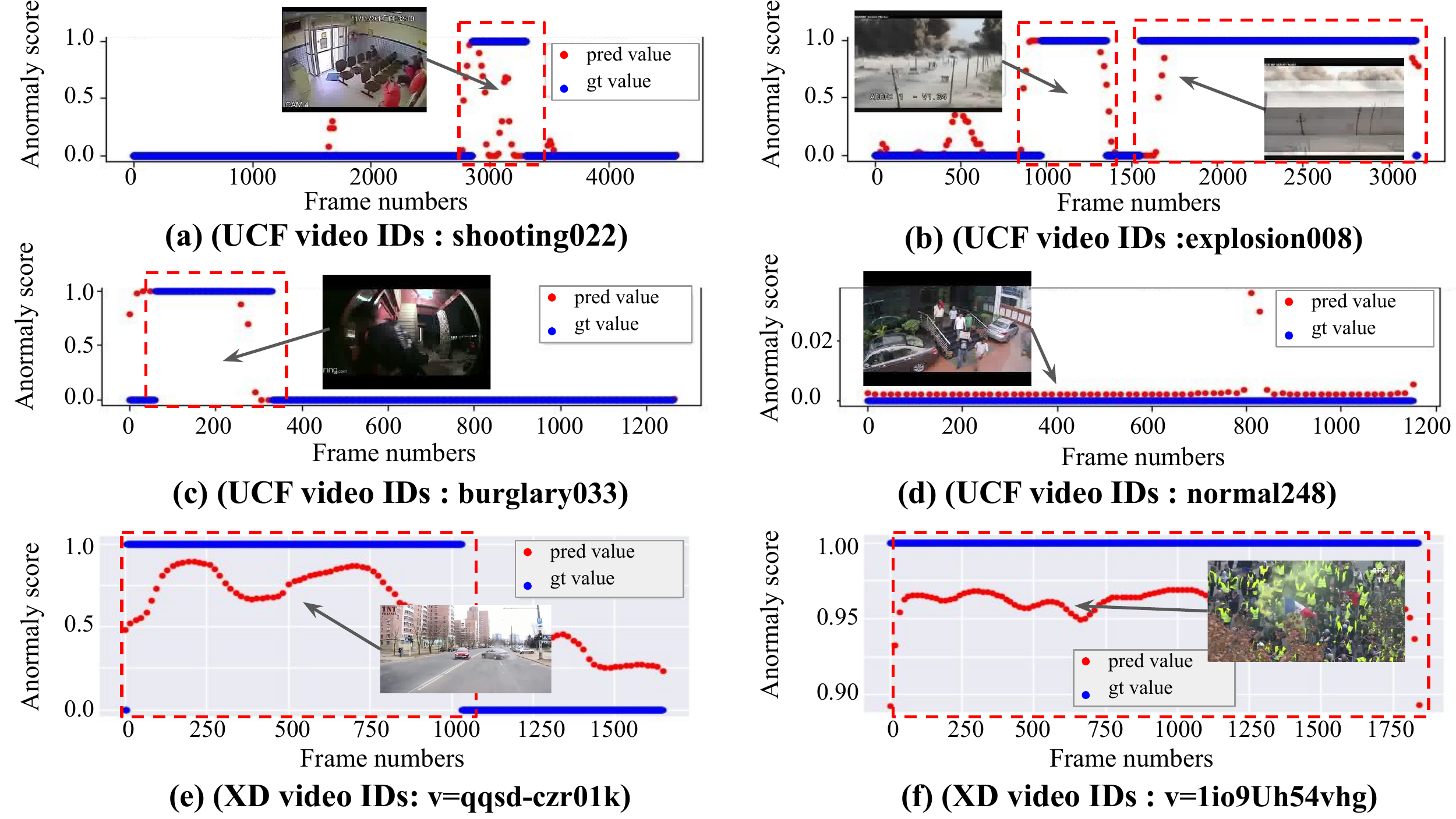}
\vspace{-5mm}
\caption{Anomaly scores (red points) predicted by our MGFN on UCF-crime (a - d), and XD-violence (e - f). Red boxes indicate the ground truth anomalies.}
\label{fig:qualitative-1}
\end{figure}

The results in Table \ref{table:backbone-design} further manifest the effectiveness of our GF mechanism. FF achieves only 82.66\% AUC on UCF datasets and 72.14\% AP on XD datasets due to the lack of global context information. When adding G in the network after learning the local information with F, the performance is boosted to 83.1$\%$ AUC on UCF and 74.30\% AP on XD, indicating the importance of the global feature. Tian \etal \cite{RTFM} and GF-Fusion
share the similar architecture of the parallel global and local feature extractor. They further improve the AUC to 84.3\% and 85.05\% respectively, indicating the complementarity of global and local features. At last, our MGFN further improves the performance to 85.80\% AUC on UCF and 78.50\% AP on XD, manifesting the effectiveness of our GF mechanism. 

\setlength{\tabcolsep}{4pt}
\begin{table}[!t]
\begin{center}
\begin{tabular}{ccc}
\hline\noalign{\smallskip}
Network structure & AUC(\%)-UCF & AP(\%)-XD \\ 
\noalign{\smallskip}
\hline
\noalign{\smallskip}
FF & 82.66 & 72.14\\
FG & 83.10 & 74.30\\
RTFM (2021) & 84.30 & 77.81\\
GF-Fusion & 85.05 & 78.03\\
\textbf{MGFN (GF)} & \textbf{85.80} & \textbf{78.50}\\
\hline
\end{tabular}
\caption{\textbf{Ablation studies of Glance-Focus mechanism} (G: Glance Block, F: Focus Block). All methods use RTFM loss \cite{RTFM} for the fair comparison. The results indicate that spatial focusing (FF), spatial focusing then glance temporal context (FG) and temporal-spatial fusion (RTFM \cite{RTFM} and GF-Fusion) network structures are all inferior than Glance-Focus mechanism (GF).}
\label{table:backbone-design}
\vspace{-3mm}
\end{center}
\end{table}
\setlength{\tabcolsep}{1.4pt}

\subsubsection{FAM and MC Loss}
In this section, we conduct ablation studies on FAM and MC loss on UCF-crime and XD-violence datasets. As shown in Table \ref{table:AB-results}, FAM enhance the model's awareness of feature magnitude, thus improving AUC by more than 1.65\% AUC on the UCF dataset and 1.02\% AP on XD dataset. Furthermore, MC loss encourages the magnitude feature learning and thus introduces nearly 2.85\% AUC improvement on UCF dataset and 3.69\% AP increment on XD dataset. In addition, combining FAM and MC Loss can further improve the performance, indicating their compatibility in video anomaly detection. 

\setlength{\tabcolsep}{4pt}
\begin{table}[!t]
\begin{center}
\begin{tabular}{ccccc}
\hline\noalign{\smallskip}
Baseline & FAM  & L$_{mc}$  & AUC(\%)-UCF & AP(\%)-XD \\
\noalign{\smallskip}
\hline
\noalign{\smallskip}
\checkmark & - & - & 83.20 & 75.11  \\
\checkmark & \checkmark  & - &  84.85 & 76.13  \\
\checkmark & - & \checkmark &  86.05 & 78.80  \\
\checkmark & \checkmark  & \checkmark &  \textbf{86.98} & \textbf{80.11}\\
\hline
\end{tabular}
\caption{\textbf{Ablation studies of FAM and MC loss} on UCF-crime and XD-Violence. ``Baseline'' means MGFN without FAM and trained only sigmoid cross-entropy loss.}
\vspace{-5mm}
\label{table:AB-results}
\end{center}
\end{table}
\setlength{\tabcolsep}{1.4pt}

\subsubsection{Qualitative Analysis}
In this section, we show the qualitative results of our MGFN. As shown in Figure \ref{fig:qualitative-1}, the distributions of red points match the ground truth very well in most cases. Specifically, our approach is able to precisely detect the abnormal events (see a,c,e), even if the video contains multiple anomalous clips (see b). For the normal video (d), our method has stable prediction throughout the video. Note that due to a sudden big movement in the video scene, the predicted abnormal probability is changed from 0 to 0.04 on the rising edge of (d), which is small enough and will not produce an alarm. It also manifests the robustness of our approach against the noise like the object movements in the video. 
For video (f) which is only composed of abnormal frames, our MGFN keeps predicting the high anomaly probability, indicating that our approach does not overfit the common scenes where most of the frames are normal. 

\section{Conclusion}
This paper introduces a novel framework MGFN with a Glance-and-Focus module and a Magnitude Contrastive loss for anomaly detection. Imitating human beings' global-to-local vision system, the proposed MGFN contains a Glance and Focus mechanism to effectively integrate the global context and local features. In addition, a Feature Amplification Mechanism (FAM) is proposed to enhance the model's awareness of feature magnitudes. To learn a scene-adaptive cross-video feature magnitude distribution, a Magnitude Contrastive loss was introduced for encouraging the separability of normal and abnormal feature magnitudes. Experimental results on two large-scale datasets UCF-Crime and XD-Voilence demonstrate that the approach outperforms the state-of-the-art works by a large margin.

\clearpage

\section{Acknowledgments}
This work was supported by the Smart Traffic Fund (PSRI/27/2201/PR) funded by the Hong Kong Productivity Council and the Transport Department of HKSAR.

\bibliographystyle{aaai23}

\end{document}